\documentclass[runningheads]{llncs}

\usepackage[T1]{fontenc}
\usepackage{graphicx}
\usepackage{amsmath,amssymb,amsfonts}
\usepackage{booktabs}
\usepackage{multirow}
\usepackage{url}
\graphicspath{{figures/}}
\usepackage[hyperfootnotes=false]{hyperref}
 \usepackage[numbers]{natbib}
\begin{document}

\title{IFW-BLS: Dual-Robust Broad Learning System with Intuitionistic Fuzzy Wave Loss}
\titlerunning{IFW-BLS}

\author{Mushir Akhtar \and M. Tanveer}
\authorrunning{M. Akhtar and M. Tanveer}
\institute{Indian Institute of Technology Indore, Simrol, Indore, India\\
\email{\{phd2101241004,mtanveer\}@iiti.ac.in}}

\maketitle

\makeatletter
\begingroup
\stepcounter{footnote}
\renewcommand{\@makefnmark}{}
\footnotetext{\textit{Preprint: Accepted at ICONIP 2026.}}
\addtocounter{footnote}{-1}
\endgroup
\makeatother

\begin{abstract}
Broad Learning System is an efficient randomized learning model that expands network width through feature and enhancement nodes and estimates the output weights without deep backpropagation. Its standard least-squares training, however, is vulnerable in two different ways: (i) large residuals caused by noise, outliers, or corrupted labels can dominate the objective, and (ii) all samples are treated as equally reliable even when some lie in ambiguous or locally conflicting regions. This paper proposes IFW-BLS, an Intuitionistic Fuzzy Wave Broad Learning System that addresses these two sources of fragility within one optimization model. The first robustness mechanism is residual-level protection, obtained by replacing the squared loss with the bounded, smooth, and asymmetric wave loss. Boundedness prevents extreme residuals from receiving unbounded influence, while asymmetry allows positive and negative deviations to be penalized differently when the dominant error direction varies. The second mechanism is sample-level credibility control, obtained through intuitionistic fuzzy scores that combine global class-center consistency with local neighborhood conflict. The resulting model evaluates the wave loss on credibility-weighted residuals, so unreliable samples are down-weighted before the bounded loss further limits the effect of extreme errors. A Nesterov accelerated gradient based optimizer is used to solve the proposed objective, avoiding the explicit matrix inversion used in conventional BLS. Experiments on UCI benchmark datasets validate the superiority of the proposed IFW-BLS model over the baseline models; additional corruption experiments also show more stable performance than BLS under noise and outlier contamination.
\keywords{Broad Learning System \and Intuitionistic Fuzzy Learning \and Wave Loss \and Robust Classification \and Randomized Neural Networks}
\end{abstract}

\section{Introduction}

Randomized neural networks have become an attractive alternative to deeply trained architectures because they reduce the number of trainable parameters and often avoid expensive backpropagation-based learning \cite{zhang2016survey,cao2018review}. Models such as random vector functional link (RVFL) networks \cite{pao1994learning,malik2023random} and extreme learning machine (ELM) \cite{huang2006extreme} randomly generate hidden transformations and learn only the output layer, which leads to fast training and simple implementation. Recent work also shows that the randomized representation itself can be made data-aware: CAWI samples input-to-hidden weights from a copula fitted to the training features, preserving inter-feature dependence while retaining closed-form output learning \cite{akhtar2026cawi}. The Broad Learning System (BLS) extends the randomized-learning idea by replacing depth with width: multiple groups of feature nodes and enhancement nodes are generated and concatenated into a broad representation \cite{chen2017broad2,chen2018universal}. This design gives BLS a favorable balance between representation ability and computational efficiency.

Despite these advantages, standard BLS remains fragile under uncertain data. Its output weights are usually obtained from a regularized least-squares problem, so large errors receive quadratic penalties. A few noisy observations, outliers, or mislabeled samples can therefore have a disproportionate effect on the learned output weights. Related robust-loss research has addressed this weakness in other learning settings. The HawkEye loss combines boundedness and smoothness with an insensitive zone for robust regression \cite{akhtar2025hawkeye}, and its integration with an RVFL network yields a robust classifier for noisy and outlier-prone data \cite{akhtar2025advancingrvfl}. In addition, the wave loss has been used in multiview learning to improve robustness while exploiting both consensus and complementary information across views \cite{quadir2025multiviewsynergy}. These results demonstrate the broader value of bounded robust losses. The conventional BLS objective, however, also does not distinguish between highly credible samples and samples located far from their class structure or near conflicting neighbors. These two issues are related but not identical. A robust loss controls the influence of large residuals after the model makes an error, whereas credibility-aware learning controls how much each sample should participate in training before its residual is penalized.

Existing BLS variants address parts of this problem from different angles. Regularized BLS improves model control through additional penalties \cite{jin2018regularized}; fuzzy BLS incorporates fuzzy subsystems for uncertainty handling \cite{feng2018fuzzy}; intuitionistic fuzzy BLS uses membership, nonmembership, and hesitation information to describe sample reliability \cite{sajid2024intuitionistic}; and robust collaborative BLS models improve learning under noisy labels through more elaborate correction or kernel-based mechanisms \cite{zheng2021broad,deng2025robust}. These methods indicate that robustness in BLS can be improved, but they also suggest a gap: residual robustness and sample credibility are often treated as separate design choices rather than as complementary components of the same training objective.

This paper develops a dual-robust BLS model from that perspective. The proposed method, named IFW-BLS for Intuitionistic Fuzzy Wave Broad Learning System, uses an intuitionistic fuzzy score to quantify the credibility of each training sample and applies the wave loss \cite{akhtar2024advancing} to the resulting score-weighted residuals. The wave loss is bounded, smooth, and asymmetric: the bound limits extreme deviations, while the asymmetry parameter shifts the relative penalty assigned to positive and negative residuals. This flexibility lets the model adapt when overestimation-like and underestimation-like deviations are not equally harmful for a dataset. The intuitionistic fuzzy score, on the other hand, uses both global class-center information and local class-conflict information to reduce the influence of ambiguous or unreliable samples. Together, these mechanisms produce robustness from two views: the credibility view decides how strongly a sample should influence learning, and the residual view decides how strongly an error should be penalized.

The main contributions are summarized as follows.
\begin{enumerate}
\item We propose IFW-BLS, an Intuitionistic Fuzzy Wave Broad Learning System that improves BLS robustness from two complementary directions: sample-level credibility modeling through intuitionistic fuzzy scores and residual-level protection through the wave loss.
\item We develop a credibility-weighted wave-loss objective in which unreliable samples are softened before the robust loss is evaluated, allowing the model to reduce the influence of ambiguous samples, noisy observations, and outliers within a unified formulation.
\item We derive the gradient of the proposed objective and optimize IFW-BLS using a Nesterov accelerated gradient based strategy, thereby avoiding the explicit matrix inversion used in conventional BLS training. 
\item We evaluate IFW-BLS on 30 UCI benchmark datasets using accuracy, average rank, and Friedman and Nemenyi statistical tests, and further examine its behavior under controlled noise and outlier contamination to verify its robustness.
\end{enumerate}

The remainder of this paper is organized as follows. Section~\ref{prelim} reviews the preliminaries of BLS, wave loss, and intuitionistic fuzzy credibility. Section~\ref{proposed-work} presents the proposed IFW-BLS model, including its objective function, gradient derivation, optimization procedure, and computational complexity. Section~\ref{experiments} reports the experimental results. Finally, Section~\ref{conclusions} concludes the paper.

\section{Preliminaries} \label{prelim}

\subsection{Notation}

Let $X=[x_1,\ldots,x_m]^\top\in\mathbb{R}^{m\times d}$ denote a training set with $m$ samples and $d$ input features. The target matrix is $Y\in\mathbb{R}^{m\times c}$, where $c$ is the number of classes and each row is a one-hot encoded label vector. For matrices, $\|\cdot\|_F$ denotes the Frobenius norm and $(\cdot)^\top$ denotes transpose. The number of feature-node windows is $q$, the number of nodes in each feature window is $p$, the number of enhancement-node windows is $s$, and the number of nodes in each enhancement window is $r$.

\subsection{Broad Learning System (BLS) \cite{chen2017broad2}}

\begin{figure}[t]
\centering
\includegraphics[width=0.8\textwidth]{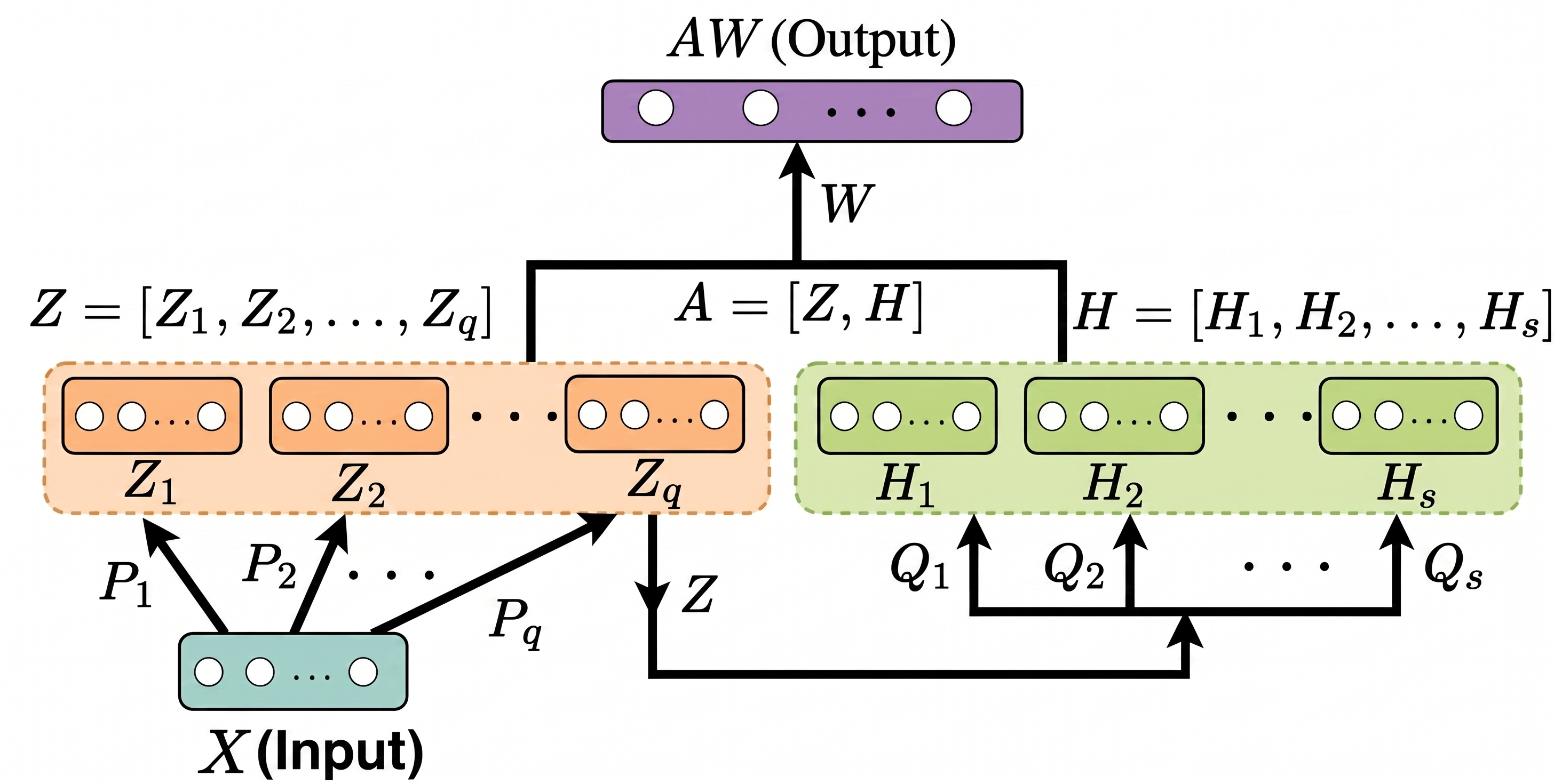}
\caption{Standard BLS architecture with randomized feature and enhancement windows.}
\label{fig:bls_architecture}
\end{figure}

BLS constructs a broad representation in two stages. Its overall information flow is shown in Fig.~\ref{fig:bls_architecture}. the input is first mapped into multiple randomized feature windows, then transformed by enhancement windows, and finally concatenated into a broad representation used to learn the output weights. Given $X$, the $k^{\mathrm{th}}$ feature window is generated as
\begin{equation}
Z_k=\phi(XP_k+\mathbf{1}b_k^\top), \qquad k=1,\ldots,q,
\end{equation}
where $P_k\in\mathbb{R}^{d\times p}$ and $b_k\in\mathbb{R}^{p}$ are randomly generated parameters, $\mathbf{1}$ is an all-one vector of length $m$, and $\phi(\cdot)$ is an activation function. The feature-node output is
\begin{equation}
Z=[Z_1,Z_2,\ldots,Z_q]\in\mathbb{R}^{m\times pq}.
\end{equation}
The enhancement layer further maps $Z$ into $s$ enhancement windows:
\begin{equation}
H_\ell=\psi(ZQ_\ell+\mathbf{1}d_\ell^\top), \qquad \ell=1,\ldots,s,
\end{equation}
where $Q_\ell\in\mathbb{R}^{pq\times r}$, $d_\ell\in\mathbb{R}^{r}$, and $\psi(\cdot)$ is an activation function. Concatenating all enhancement windows gives
\begin{equation}
H=[H_1,H_2,\ldots,H_s]\in\mathbb{R}^{m\times rs}.
\end{equation}
The final broad representation is
\begin{equation}
A=[Z,H]\in\mathbb{R}^{m\times n_b}, \qquad n_b=pq+rs.
\end{equation}

The conventional BLS output matrix $W\in\mathbb{R}^{n_b\times c}$ is estimated by
\begin{equation}\label{eq:bls_ls}
\min_W \; \frac{1}{2}\|W\|_F^2+\frac{C}{2}\|AW-Y\|_F^2,
\end{equation}
where $C>0$ is the regularization parameter. Problem \eqref{eq:bls_ls} has a closed-form solution involving an inverse of either $A^\top A+C^{-1}I$ or $AA^\top+C^{-1}I$. This is efficient for moderate sizes, but it becomes expensive for very broad representations or large training sets and remains sensitive to extreme residuals.

\subsection{Wave Loss \cite{akhtar2024advancing}}

\begin{figure}[t]
\centering
\begin{minipage}[t]{0.32\textwidth}
\centering
\includegraphics[width=\linewidth]{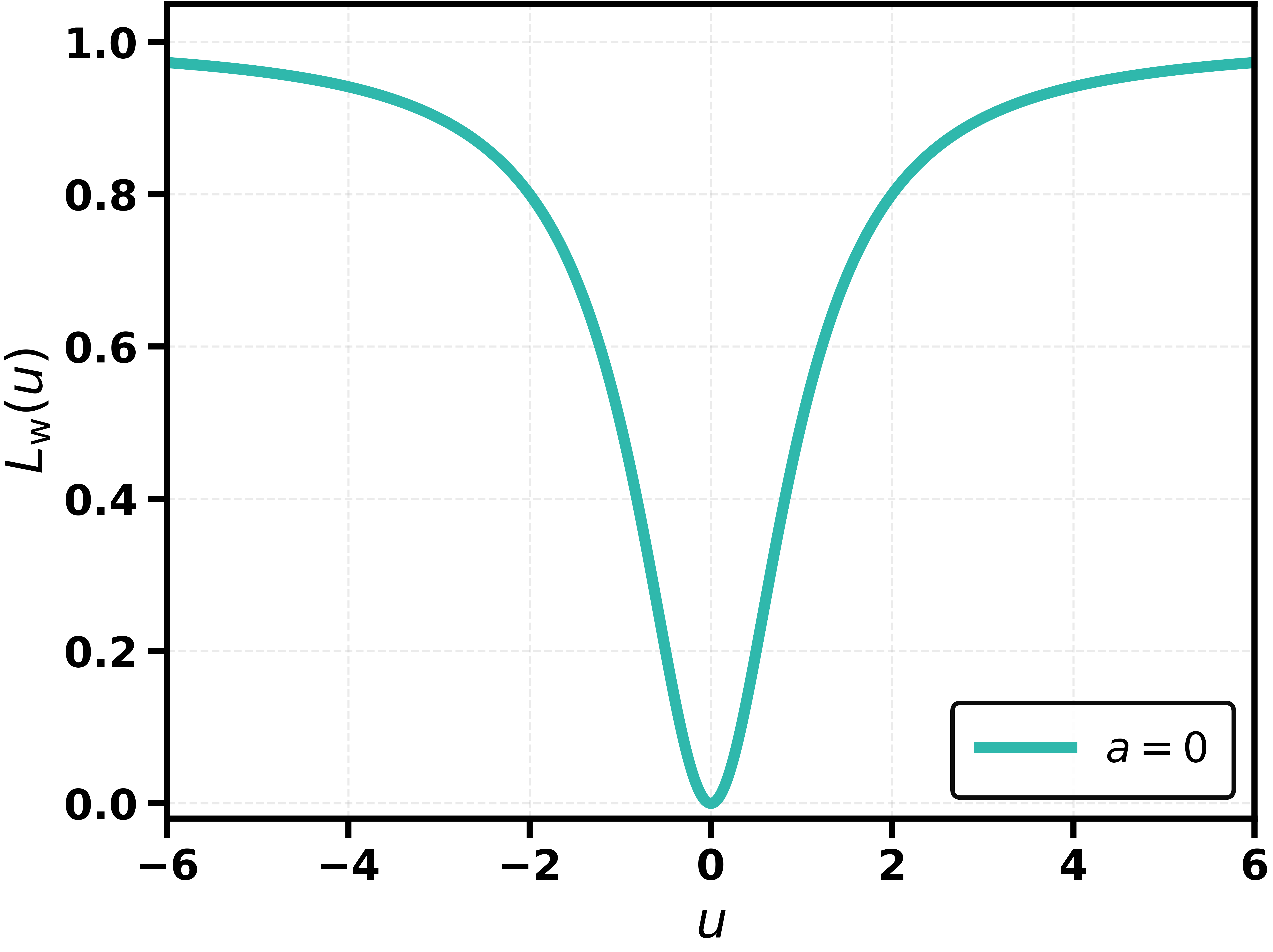}\\[1mm]
\textbf{(a)} $a=0$
\end{minipage}\hfill
\begin{minipage}[t]{0.32\textwidth}
\centering
\includegraphics[width=\linewidth]{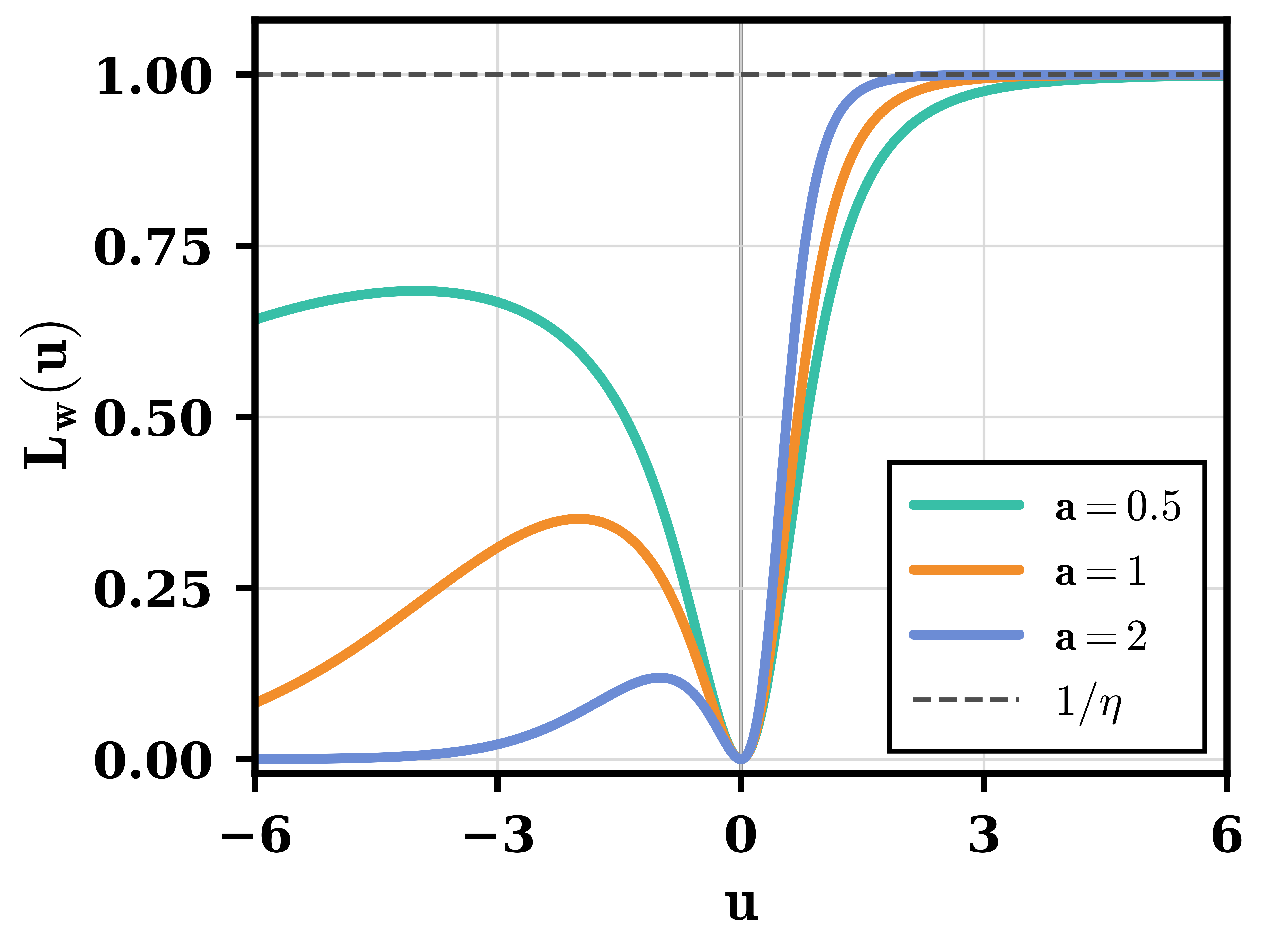}\\[1mm]
\textbf{(b)} $a>0$
\end{minipage}\hfill
\begin{minipage}[t]{0.32\textwidth}
\centering
\includegraphics[width=\linewidth]{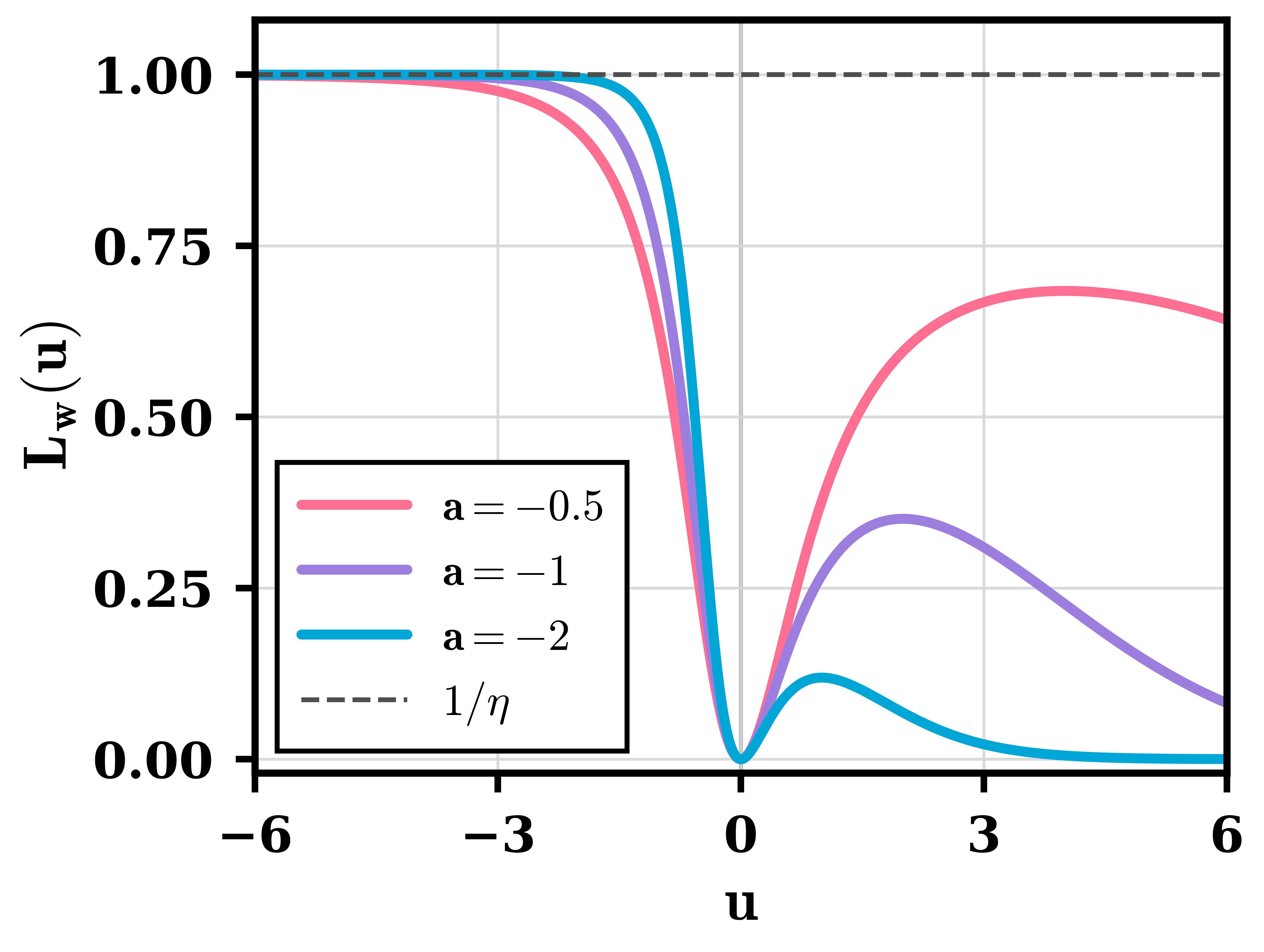}\\[1mm]
\textbf{(c)} $a<0$
\end{minipage}
\caption{Wave-loss curves under fixed bound parameter $\eta=1$: symmetric, positive-asymmetric, and negative-asymmetric cases.}
\label{fig:wave_loss_curves}
\end{figure}

For a residual $u\in\mathbb{R}$, the wave loss is defined as 
\begin{equation}\label{eq:wave}
L_{\mathrm{w}}(u)=
\frac{u^2\exp(a u)}
{1+\lambda u^2\exp(a u)},
\end{equation}
where $a\in\mathbb{R}$ controls the asymmetry and $\lambda>0$ controls the upper bound. Since $u^2\exp(au)\ge 0$, the loss satisfies $0\le L_{\mathrm{w}}(u)<1/\lambda$. Thus, very large residuals cannot grow without limit in the objective. The sign and magnitude of $a$ also move the emphasis of the loss: $a=0$ gives a symmetric penalty, whereas $a>0$ and $a<0$ assign different penalties to positive and negative residuals. The derivative needed for optimization is
\begin{equation}\label{eq:wave_derivative}
L_{\mathrm{w}}'(u)=
\frac{u(2+a u)\exp(a u)}
{\left(1+\lambda u^2\exp(a u)\right)^2}.
\end{equation}
This smooth derivative allows the loss to be used in gradient-based BLS training. Fig.~\ref{fig:wave_loss_curves} illustrates how $a=0$ gives a symmetric curve, while positive and negative values of $a$ shift the dominant penalization direction.

\subsection{Intuitionistic Fuzzy Credibility}

Intuitionistic fuzzy theory represents uncertainty through membership, nonmembership, and hesitation degrees \cite{zadeh1965fuzzy,atanassov1999intuitionistic}. Following the kernelized credibility modeling used in intuitionistic fuzzy classifiers \cite{rezvani2019intuitionistic}, each sample receives a score that reflects two reliability cues: how close it is to the center of its own class and how much local class conflict surrounds it.

Let $\varphi(\cdot)$ denote the feature map induced by an RBF kernel
\begin{equation}
K(x_i,x_j)=\exp(-\gamma_k\|x_i-x_j\|_2^2),
\end{equation}
where $\gamma_k>0$. Distances in the induced feature space are computed by
\begin{equation}\label{eq:kernel_distance}
\|\varphi(x_i)-\varphi(x_j)\|_2 =
\sqrt{K(x_i,x_i)+K(x_j,x_j)-2K(x_i,x_j)}.
\end{equation}
For class $g\in\{1,\ldots,c\}$, define its feature-space center and radius as
\begin{equation}
D_g=\frac{1}{m_g}\sum_{y_i=g}\varphi(x_i), \qquad
R_g=\max_{y_i=g}\|\varphi(x_i)-D_g\|_2,
\end{equation}
where $m_g$ is the number of samples in class $g$. For sample $x_i$ with label $y_i$, the membership degree is
\begin{equation}\label{eq:membership}
\mu_i=\max\left\{0,\;1-\frac{\|\varphi(x_i)-D_{y_i}\|_2}{R_{y_i}+\varepsilon}\right\},
\end{equation}
where $\varepsilon>0$ avoids division by zero. The required class-center distance can be computed only through kernel evaluations as
\begin{align}
\|\varphi(x_i)-D_g\|_2^2
&=K(x_i,x_i)
-\frac{2}{m_g}\sum_{y_k=g}K(x_i,x_k)
+
\frac{1}{m_g^2}\sum_{y_k=g}\sum_{y_\ell=g}K(x_k,x_\ell).
\end{align}
Thus, the intuitionistic fuzzy scores do not require explicit construction of $\varphi(\cdot)$.
The local conflict ratio is
\begin{equation}
\Psi_i=
\frac{
\left|\{x_j\in\mathcal{N}_\tau(i): y_j\ne y_i\}\right|
}{
\max\{1,|\mathcal{N}_\tau(i)|\}
},
\end{equation}
where $\mathcal{N}_\tau(i)=\{x_j:\|\varphi(x_i)-\varphi(x_j)\|_2\le\tau,\; j\ne i\}$ and $\tau>0$ is a neighborhood radius. The nonmembership degree is then
\begin{equation}\label{eq:nonmembership}
\nu_i=(1-\mu_i)\Psi_i.
\end{equation}
This construction gives high nonmembership only when a sample is globally weak for its own class and locally surrounded by conflicting labels. The remaining hesitation is $\pi_i=1-\mu_i-\nu_i$. Finally, the credibility score is
\begin{equation}\label{eq:score}
s_i=
\begin{cases}
\mu_i, & \nu_i=0,\\[2mm]
0, & \mu_i\le \nu_i,\\[2mm]
\dfrac{1-\nu_i}{2-\mu_i-\nu_i}, & \text{otherwise},
\end{cases}
\end{equation}
with $s_i\in[0,1]$. Larger $s_i$ means that the sample is more reliable for training; smaller $s_i$ reduces its influence.

\section{Proposed Dual-Robust BLS}\label{proposed-work}

\subsection{Motivation}

A single robustness mechanism is often insufficient. If only a bounded loss is used, a highly suspicious sample may still influence the model until its residual becomes large enough to be saturated. If only a credibility score is used, moderate but systematic residuals may still be penalized by a squared loss. IFW-BLS therefore combines the two mechanisms directly: sample credibility scales the residual first, and the wave loss then evaluates the scaled residual with a bounded and direction-sensitive penalty.



\subsection{Credibility-Weighted Wave Loss}

Let $e_{ij}=(AW)_{ij}-Y_{ij}$ be the residual of sample $i$ for output class $j$. The proposed credibility-weighted residual is
\begin{equation}\label{eq:weighted_residual}
\xi_{ij}=s_i e_{ij}=s_i\left((AW)_{ij}-Y_{ij}\right).
\end{equation}
The proposed IFW-BLS objective is
\begin{equation}\label{eq:main_objective}
\min_W J(W)=
\frac{1}{2}\|W\|_F^2+
C\sum_{i=1}^{m}\sum_{j=1}^{c} L_{\mathrm{w}}(\xi_{ij}).
\end{equation}
The role of $s_i$ in \eqref{eq:main_objective} is not merely a post-hoc sample weight. It changes the argument of the robust loss itself. Consequently, low-credibility samples are softened before the wave loss is evaluated, and samples with $s_i=0$ contribute no loss or gradient. At the same time, the wave loss caps the contribution of large weighted residuals and uses $a$ to control whether positive or negative deviations should receive stronger penalization.

For $\lambda>0$, the wave loss satisfies
\begin{equation}
0\le L_{\mathrm{w}}(u)=\frac{u^2\exp(a u)}{1+\lambda u^2\exp(a u)}<\frac{1}{\lambda}.
\end{equation}
With the weighted residual $\xi_{ij}=s_i((AW)_{ij}-Y_{ij})$ and $s_i\in[0,1]$, each loss term is also bounded:
\begin{equation}
0\le L_{\mathrm{w}}(\xi_{ij})<\frac{1}{\lambda}.
\end{equation}
Moreover, if $s_i=0$, then $\xi_{ij}=0$ and the sample contributes neither loss nor gradient. For $0<s_i<1$, the residual entering the loss is contracted, and the gradient contribution receives an additional multiplicative factor $s_i$ from the chain rule. The parameter $a$ controls how positive and negative residuals are penalized relative to each other, so IFW-BLS can adapt the direction of stronger penalization.

\subsection{Gradient Derivation}

Using \eqref{eq:wave_derivative} and the chain rule, the contribution of sample $i$ to the gradient for output class $j$ is
\begin{equation}\label{eq:scalar_grad}
\Delta_{ij}=
s_i\,
\frac{\xi_{ij}(2+a\xi_{ij})\exp(a\xi_{ij})}
{\left(1+\lambda \xi_{ij}^{2}\exp(a\xi_{ij})\right)^2}.
\end{equation}
Therefore, for the $j^{\mathrm{th}}$ column $W_{:j}$ of $W$,
\begin{equation}\label{eq:class_grad}
\nabla_{W_{:j}}J(W)=
W_{:j}+C\sum_{i=1}^{m}\Delta_{ij}A_{i:}^{\top}.
\end{equation}
Equivalently, if $\Delta\in\mathbb{R}^{m\times c}$ collects all $\Delta_{ij}$ terms, the full gradient can be written compactly as
\begin{equation}\label{eq:full_grad}
\nabla J(W)=W+C A^\top \Delta.
\end{equation}
The derivation shows the two damping effects explicitly: $s_i$ appears in the residual $\xi_{ij}$ and again through the derivative $\partial \xi_{ij}/\partial W_{:j}=s_i A_{i:}^{\top}$.

\subsection{Optimization}

The proposed objective is smooth for fixed IF scores and fixed randomized BLS features. We optimize \eqref{eq:main_objective} using Nesterov accelerated gradient (NAG)-based algorithm as used in \cite{10685140,AKHTAR2026112711}, which avoids the matrix inverse in the conventional BLS solution and is suitable for broad representations. Let $V^{(t)}$ denote the velocity matrix at iteration $t$, and let $\eta_0$, $\rho$, and $\gamma$ denote the initial learning rate, learning-rate decay, and momentum coefficient, respectively. The optimization proceeds as follows.

\begin{enumerate}
    \item \textbf{Initialization.} Construct the BLS representation $A$, compute the credibility scores $\{s_i\}_{i=1}^{m}$, initialize the output weights $W^{(0)}$, set $V^{(0)}=\mathbf{0}$, and choose $C$, $a$, $\lambda$, $\eta_0$, $\rho$, $\gamma$, tolerance $\delta$, and maximum iteration number $T$.

    \item \textbf{Look-ahead point.} At iteration $t$, NAG first evaluates the objective at the momentum-aided point
    \begin{equation}
    \widehat{W}^{(t)}=W^{(t)}+\gamma V^{(t)}.
    \end{equation}
    This look-ahead step uses the current search direction before computing the gradient, which usually gives more stable updates than plain gradient descent.

    \item \textbf{Credibility-weighted residuals.} Using $\widehat{W}^{(t)}$, compute
    \begin{equation}
    \xi_{ij}^{(t)}=s_i\left((A\widehat{W}^{(t)})_{ij}-Y_{ij}\right), \qquad i=1,\ldots,m,\; j=1,\ldots,c.
    \end{equation}
    The IF score enters before the wave loss is evaluated, so samples with weak credibility have reduced residual contributions.

    \item \textbf{Gradient evaluation.} For all samples and classes, compute
    \begin{equation}
    \Delta_{ij}^{(t)}=
    s_i\,
    \frac{\xi_{ij}^{(t)}(2+a\xi_{ij}^{(t)})\exp(a\xi_{ij}^{(t)})}
    {\left(1+\lambda (\xi_{ij}^{(t)})^{2}\exp(a\xi_{ij}^{(t)})\right)^2},
    \end{equation}
    and form the full gradient
    \begin{equation}
    G^{(t)}=\widehat{W}^{(t)}+C A^\top\Delta^{(t)}.
    \end{equation}

    \item \textbf{Parameter update.} The learning rate is decayed as
    \begin{equation}
    \eta_t=\eta_0\exp(-\rho t).
    \end{equation}
    The velocity and output weights are then updated by
    \begin{equation}
    V^{(t+1)}=\gamma V^{(t)}-\eta_tG^{(t)}, \qquad
    W^{(t+1)}=W^{(t)}+V^{(t+1)}.
    \end{equation}
    The iterations stop when $t\ge T$ or $\|W^{(t+1)}-W^{(t)}\|_F\le\delta$.
\end{enumerate}

For a test sample $\tilde{x}$, its BLS representation $\tilde{a}$ is generated using the same fixed feature and enhancement mappings as the training data. The predicted class is
\begin{equation}
\hat{y}=\arg\max_{j\in\{1,\ldots,c\}}(\tilde{a}W)_j.
\end{equation}


\subsection{Computational Complexity}

Let $n_b=pq+rs$ be the width of the final BLS representation. Constructing the randomized feature windows requires $\mathcal{O}(mdpq)$ operations, and constructing the enhancement windows requires $\mathcal{O}(mpqrs)$ operations when each enhancement window is generated from the full feature-node matrix. The intuitionistic fuzzy scores are computed once before optimization. With an RBF kernel, a direct implementation forms the $m\times m$ kernel matrix in $\mathcal{O}(m^2d)$ time and then computes class-center distances and neighborhood conflicts from this matrix. After $A$ and $\{s_i\}_{i=1}^{m}$ are fixed, each NAG iteration is dominated by the products $A\widehat{W}$ and $A^\top\Delta$, giving a cost of $\mathcal{O}(mn_bc)$. The element-wise evaluation of the wave-loss derivative costs $\mathcal{O}(mc)$, and the velocity and weight updates cost $\mathcal{O}(n_bc)$. Thus, the total optimization cost over $T$ iterations is $\mathcal{O}(Tmn_bc).$ The overall training cost of IFW-BLS is therefore
$\mathcal{O}(mdpq+mpqrs+m^2d+Tmn_bc),$
excluding lower-order terms. Conventional BLS avoids iterative optimization but requires a matrix inverse with cost $\mathcal{O}(\min\{n_b^3,m^3\})$ after the broad representation is constructed. IFW-BLS replaces this inversion by first-order updates while adding a one-time credibility computation, which is appropriate when robustness to sample ambiguity and residual outliers is required.

\section{Experiments and Discussion}
\label{experiments}

In this section, we evaluate the proposed IFW-BLS model on 30 UCI benchmark datasets against RVFL \cite{pao1994learning}, RVFL without direct link (RVFLwoDL/ELM) \cite{huang2006extreme}, BLS \cite{chen2017broad2}, Wave-RVFL \cite{sajid2024wavervflrandomizedneuralnetwork}, NF-BLS \cite{feng2018fuzzy}, F-BLS \cite{sajid2024intuitionistic}, IF-BLS \cite{sajid2024intuitionistic}, and KRP-BLS \cite{deng2025robust}. The methods are compared using classification accuracy, average rank, and Friedman and Nemenyi statistical tests. In addition, IFW-BLS is evaluated under controlled noise and outlier contamination to examine its robustness on corrupted training data.

\subsection{Experimental Setup and Hyperparameter Selection}

All experiments are implemented in MATLAB R2023a on a Windows~11 workstation with an 11th-generation Intel Core i7-11700 CPU at 2.50~GHz and 16~GB RAM. For a fair comparison, all methods are evaluated using 5-fold cross-validation together with grid-based hyperparameter selection. In each run, the dataset is divided into five disjoint folds; four folds are used for training and the remaining fold is used for testing. The final performance of a model is reported as the best mean testing accuracy obtained over the considered hyperparameter grid.

The regularization parameter \(C\) is searched over \(\{10^{-6},10^{-4},\ldots,10^{6}\}\) for all methods. For RVFL and RVFLwoDL, the number of hidden nodes is selected from \(5{:}10{:}205\). For BLS-based models, including the proposed IFW-BLS, the number of feature-node windows is chosen from \(1{:}2{:}21\), the number of feature nodes in each window from \(5{:}5{:}50\), and the number of enhancement nodes from \(5{:}10{:}105\). For NF-BLS, the number of fuzzy groups is tuned over \(1{:}2{:}21\), the number of fuzzy nodes per group over \(5{:}5{:}50\), and the number of enhancement nodes over \(5{:}10{:}105\). The hyperparameter settings of Wave-RVFL follow \cite{sajid2024wavervflrandomizedneuralnetwork}; those of F-BLS and IF-BLS follow \cite{sajid2024intuitionistic}; and KRP-BLS is configured according to its original implementation \cite{deng2025robust}. For the proposed IFW-BLS, the wave-loss asymmetry parameter is selected from $a\in\{-3,-4,\ldots,3\}$ and the bound parameter from $\lambda\in\{0.1,0.5,1\}$. The RBF kernel width $\gamma_k$ used in the intuitionistic fuzzy score is selected from $\{10^{-6},10^{-4},\ldots,10^{6}\}$. The NAG parameters are fixed as $W^{(0)}=0.01\mathbf{1}$, $V^{(0)}=\mathbf{0}$, initial learning rate $\eta_0=0.01$, decay $\rho=0.1$, momentum $\gamma=0.6$, tolerance $\delta=10^{-6}$, and maximum iterations $T=100$.

\subsection{Performance Evaluation}

Table~\ref{tab:avg_results} reports the average accuracy and average rank of all compared models over the 30 UCI datasets. In terms of average accuracy, the proposed IFW-BLS obtains the best result of $87.4896\%$. The closest competitor is IF-BLS with $83.9102\%$, followed by KRP-BLS with $83.002\%$, standard BLS with $82.0751\%$, F-BLS with $82.0376\%$, Wave-RVFL with $81.5155\%$, NF-BLS with $80.0399\%$, RVFL with $79.8876\%$, and RVFLwoDL with $79.4376\%$. Thus, IFW-BLS improves the average accuracy by $3.5794\%$ over IF-BLS, $4.4876\%$ over KRP-BLS, $5.4145\%$ over standard BLS, $5.452\%$ over F-BLS, $5.9741\%$ over Wave-RVFL, $7.4497\%$ over NF-BLS, $7.602\%$ over RVFL, and $8.052\%$ over RVFLwoDL. These accuracy results support the dual-robustness motivation of IFW-BLS. The strong performance of IF-BLS confirms that intuitionistic fuzzy sample credibility is useful for reducing the influence of ambiguous or unreliable samples. However, IFW-BLS further improves over IF-BLS by applying the bounded and asymmetric wave loss to the credibility-weighted residuals. This indicates that sample-level reliability alone is not sufficient; after low-credibility samples are softened, residual-level protection is still beneficial for controlling large deviations and adapting the penalty to the dominant error direction. The improvement over Wave-RVFL also shows that wave-loss robustness becomes more effective when it is embedded in the BLS feature-enhancement representation and combined with intuitionistic fuzzy credibility.

\begin{table}[t]
\centering
\scriptsize
\caption{Performance comparison on 30 UCI datasets. Best and second-best values are shown in bold and underlined, respectively. $\dagger$ denotes the proposed model.}
\label{tab:avg_results}
\setlength{\tabcolsep}{1.6pt}
\resizebox{\textwidth}{!}{%
\begin{tabular}{lccccccccc}
\toprule
\textbf{Dataset} & \textbf{RVFL}~\cite{pao1994learning} & \textbf{RVFLwoDL}~\cite{huang2006extreme} & \textbf{BLS}~\cite{chen2017broad2} & \textbf{Wave-RVFL}~\cite{sajid2024wavervflrandomizedneuralnetwork} & \textbf{NF-BLS}~\cite{feng2018fuzzy} & \textbf{F-BLS}~\cite{sajid2024intuitionistic} & \textbf{IF-BLS}~\cite{sajid2024intuitionistic} & \textbf{KRP-BLS}~\cite{deng2025robust} & \textbf{IFW-BLS}$^{\dagger}$ \\
\midrule
acute\_inflammation & \textbf{100} & \textbf{100} & \textbf{100} & \textbf{100} & \textbf{100} & \textbf{100} & \textbf{100} & \textbf{100} & \textbf{100} \\
acute\_nephritis & \textbf{100} & \textbf{100} & \textbf{100} & \textbf{100} & \textbf{100} & \textbf{100} & \textbf{100} & \textbf{100} & \textbf{100} \\
bank & \textbf{90.9051} & 89.4051 & 89.7366 & 88.1779 & 89.5817 & 89.759 & 89.4051 & 87.723 & \underline{90.561} \\
blood & 78.4056 & 76.9056 & 79.6272 & \underline{80.7578} & 76.0807 & 78.5807 & 77.2451 & 76.7 & \textbf{83.5711} \\
breast\_cancer & 68.1727 & 66.6727 & 69.8851 & 70.2179 & 70.1754 & 72.2868 & \underline{83.1579} & 82.4947 & \textbf{91.0637} \\
breast\_cancer\_wisc & 89.4897 & 87.9897 & 88.4183 & 86.805 & \underline{90.7081} & 88.2775 & 88.9938 & 89.324 & \textbf{98.4588} \\
chess\_krvkp & 73.5312 & 72.0312 & 84.3862 & 75.7371 & 70.4004 & 84.2921 & 84.9184 & \underline{87.3718} & \textbf{89.9404} \\
conn\_bench\_sonar\_mines\_rocks & 62.0807 & 60.5807 & 69.2451 & 63.9431 & 60.6272 & 69.1521 & \underline{80.2091} & 78.2049 & \textbf{94.5878} \\
credit\_approval & 86.8623 & 85.3623 & 87.5362 & 88.5995 & 84.4928 & 86.3768 & 88.5507 & \textbf{89.7797} & \underline{88.7882} \\
cylinder\_bands & 67.7193 & 66.2193 & 69.9258 & 69.7509 & 69.5336 & 69.1338 & \textbf{72.8536} & \underline{70.8109} & 70.0391 \\
echocardiogram & 85.4031 & 83.9031 & 83.9316 & \underline{87.9652} & 80.9402 & 84.6724 & \textbf{88.49} & 86.7202 & 87.0762 \\
fertility & 89.5 & 90 & 90 & 86.815 & \underline{92} & 91 & 91 & 89.18 & \textbf{92.4064} \\
haberman\_survival & 74.9902 & 73.4902 & 70.275 & \underline{77.2399} & 73.4902 & 69.6563 & 75.4574 & 76.1937 & \textbf{79.2877} \\
hepatitis & 83.2258 & 83.2258 & 85.1613 & 85.7226 & \textbf{87.7419} & 84.5161 & \textbf{87.7419} & 85.1096 & \underline{86.4488} \\
hill\_valley & 77.9706 & 77.9706 & \textbf{82.0185} & 80.3097 & 78.9583 & \underline{81.6002} & 79.6249 & 78.6344 & 80.2332 \\
horse\_colic & 84.2098 & 84.7098 & 86.1422 & \underline{86.7361} & 83.9726 & 86.1385 & 86.6938 & 84.092 & \textbf{87.1719} \\
mammographic & 79.0889 & 79.0889 & 78.6712 & \underline{81.4616} & 79.504 & 78.6744 & 79.8192 & 79.8238 & \textbf{87.0013} \\
molec\_biol\_promoter & 68.961 & 68.961 & 83.9394 & 71.0298 & 75.1515 & 84.9351 & 88.7879 & \underline{89.1241} & \textbf{96.7207} \\
monks\_1 & 83.2497 & 83.2497 & 75.3314 & \underline{85.7472} & \textbf{86.1277} & 77.1396 & 77.6705 & 74.0404 & 75.5649 \\
musk\_1 & 67.864 & 67.864 & 75.8311 & 69.8999 & 67.8487 & 76.6754 & 78.9846 & \underline{79.8523} & \textbf{97.3361} \\
oocytes\_merluccius\_nucleus\_4d & 79.8407 & 79.8407 & \textbf{82.7776} & \underline{82.2359} & 80.142 & 81.8011 & 80.9201 & 79.3017 & 75.4261 \\
oocytes\_trisopterus\_nucleus\_2f & 76.4211 & 76.4211 & \textbf{78.9371} & \underline{78.7137} & 75.9911 & 78.6123 & 75.9899 & 74.2302 & 75.0727 \\
pima & 71.4888 & 71.4888 & 72.0958 & 73.6335 & 72.7918 & 71.4846 & 73.3079 & \underline{74.1087} & \textbf{78.1658} \\
pittsburg\_bridges\_T\_OR\_D & 87.6429 & 88.1429 & 89.1095 & \underline{90.2722} & 90.1429 & 88.1905 & 90.2381 & 87.0791 & \textbf{93.3928} \\
spambase & 87.1125 & 87.1125 & 89.0201 & 89.7259 & 83.2657 & 89.4582 & \underline{90.7407} & 87.4741 & \textbf{94.0153} \\
spect & 67.9245 & 67.9245 & 69.434 & 69.9622 & 67.9245 & 69.434 & \underline{72.4528} & 72.3517 & \textbf{72.9247} \\
statlog\_heart & 80 & 80 & 82.2222 & 82.4 & 80.7407 & 82.2222 & 84.0741 & \underline{85.0566} & \textbf{88.244} \\
tic\_tac\_toe & 86.0068 & 86.0068 & \underline{98.4315} & 88.587 & 82.7623 & 97.8081 & 97.0768 & 94.1645 & \textbf{99.5474} \\
titanic & 77.9168 & 77.9168 & 77.9168 & \underline{80.2543} & 78.0537 & 77.9168 & 79.9532 & 78.3541 & \textbf{95.2742} \\
vertebral column 2clases & 70.6452 & 70.6452 & 72.2453 & 72.7646 & 72.0465 & 71.3333 & \underline{72.9479} & 72.759 & \textbf{76.3667} \\
\midrule
\textbf{Average accuracy} & 79.8876 & 79.4376 & 82.0751 & 81.5155 & 80.0399 & 82.0376 & \underline{83.9102} & 83.002 & \textbf{87.4896} \\
\textbf{Average rank} & 6.5333 & 7.2167 & 4.9333 & 4.1333 & 6.1 & 5.3333 & \underline{3.55} & 4.8667 & \textbf{2.3333} \\
\bottomrule
\end{tabular}}
\end{table}

Although average accuracy provides a direct summary of predictive performance, it may be affected by large gains on a small number of datasets. Therefore, average rank is also reported to assess the consistency of each method across datasets. A lower rank indicates that a model more frequently appears among the top-performing methods. IFW-BLS achieves the best average rank of $2.3333$, followed by IF-BLS with $3.55$, Wave-RVFL with $4.1333$, KRP-BLS with $4.8667$, standard BLS with $4.9333$, F-BLS with $5.3333$, NF-BLS with $6.1$, RVFL with $6.5333$, and RVFLwoDL with $7.2167$. Compared with IFW-BLS, the rank gaps are $1.2167$ for IF-BLS, $1.8$ for Wave-RVFL, $2.5334$ for KRP-BLS, $2.6$ for standard BLS, $3$ for F-BLS, $3.7667$ for NF-BLS, $4.2$ for RVFL, and $4.8834$ for RVFLwoDL. The rank analysis further confirms that the superiority of IFW-BLS is not due to isolated improvements on only a few datasets. IFW-BLS obtains the best or tied-best result on 21 out of 30 datasets and the second-best result on 3 additional datasets. It also wins or ties against IF-BLS on 24 datasets and against standard BLS on 27 datasets. These observations show that the proposed combination of intuitionistic fuzzy credibility and wave-loss residual robustness provides stable gains across heterogeneous classification tasks.

\subsection{Statistical Significance}

Following the statistical comparison protocol recommended by Demsar \cite{demvsar2006statistical}, we use the nonparametric Friedman test to examine whether the differences in average ranks are statistically meaningful across all datasets. Let $\mathcal{K}$ be the number of compared models and $\mathcal{N}$ be the number of datasets. The null hypothesis states that all $\mathcal{K}$ models have equivalent predictive performance, so their rank differences over $\mathcal{N}$ datasets are due only to random variation. The Friedman statistic is computed as
\begin{equation}
\chi_F^2=
\frac{12\mathcal{N}}{\mathcal{K}(\mathcal{K}+1)}
\left[
\sum_{k=1}^{\mathcal{K}}\varrho(k,\bullet)^2
-
\frac{\mathcal{K}(\mathcal{K}+1)^2}{4}
\right],
\end{equation}
where $\varrho(k,\bullet)$ denotes the average rank of the $k^{\mathrm{th}}$ model. Since the Friedman statistic can be conservative, we also report the Iman-Davenport corrected statistic
\begin{equation}
F_F=
\chi_F^2
\left(
\frac{\mathcal{N}-1}
{\mathcal{N}(\mathcal{K}-1)-\chi_F^2}
\right),
\end{equation}
which follows an $F$ distribution with $(\mathcal{K}-1)$ and $(\mathcal{K}-1)(\mathcal{N}-1)$ degrees of freedom. In this study, $\mathcal{K}=9$ methods are compared on $\mathcal{N}=30$ datasets. As shown in Table~\ref{tab:statistical_tests}, the Friedman statistic is $\chi_F^2=74.2911$, and the corrected statistic is $F_F=13.0014$. The critical value at the $5\%$ significance level is $F(8,232)=1.9784$. Since $13.0014>1.9784$, the null hypothesis is rejected, indicating that the observed rank differences among the competing models are statistically significant.

After rejecting the null hypothesis, we apply the Nemenyi post-hoc test to compare IFW-BLS with each competing method. Two models are considered significantly different when the absolute difference between their average ranks is larger than the critical difference
\begin{equation}
\mathrm{C.D.}=
q_{\alpha}
\sqrt{\frac{\mathcal{K}(\mathcal{K}+1)}{6\mathcal{N}}},
\end{equation}
where $q_{\alpha}$ is the critical value for the two-tailed Nemenyi test. At $\alpha=0.10$, the critical difference is $\mathrm{C.D.}=2.0188$ for $\mathcal{K}=9$ and $\mathcal{N}=30$.

\begin{table}[t]
\centering
\caption{Statistical comparison based on Friedman and Nemenyi tests.}
\label{tab:statistical_tests}
\resizebox{\textwidth}{!}{%
\begin{tabular}{lccc}
\toprule
\multicolumn{4}{c}{Friedman test with Iman-Davenport correction} \\
\midrule
$\mathcal{K}$ & $\mathcal{N}$ & $\chi_F^2$ / $F_F$ & Critical value at $\alpha=0.05$ \\
9 & 30 & 74.2911 / 13.0014 & $F(8,232)=1.9784$ \\
\midrule
\multicolumn{4}{c}{Nemenyi post-hoc test against IFW-BLS, C.D.$=2.0188$} \\
\midrule
Model & Average rank & Rank difference & Significant \\
\midrule
RVFL~\cite{pao1994learning} & 6.5333 & 4.2 & Yes \\
RVFLwoDL~\cite{huang2006extreme} & 7.2167 & 4.8833 & Yes \\
BLS~\cite{chen2017broad2} & 4.9333 & 2.6 & Yes \\
Wave-RVFL~\cite{sajid2024wavervflrandomizedneuralnetwork} & 4.1333 & 1.8 & No \\
NF-BLS~\cite{feng2018fuzzy} & 6.1 & 3.7667 & Yes \\
F-BLS~\cite{sajid2024intuitionistic} & 5.3333 & 3 & Yes \\
IF-BLS~\cite{sajid2024intuitionistic} & 3.55 & 1.2167 & No \\
KRP-BLS~\cite{deng2025robust} & 4.8667 & 2.5333 & Yes \\
\bottomrule
\end{tabular}}
\end{table}

The Nemenyi test shows that IFW-BLS is significantly better than RVFL, RVFLwoDL, BLS, NF-BLS, F-BLS, and KRP-BLS. The differences with Wave-RVFL and IF-BLS do not exceed the critical difference, which is reasonable since these are the two baselines most closely related to the proposed design: Wave-RVFL uses wave-loss-based residual robustness, whereas IF-BLS uses intuitionistic fuzzy credibility. IFW-BLS still obtains the best average rank and average accuracy among all methods, indicating that combining these two robustness views inside BLS gives the strongest overall performance while remaining statistically competitive with the closest robust alternatives.

\subsection{Robustness Analysis under Noise and Outliers}

To directly examine the dual-robustness claim under adverse training conditions, we conduct controlled corruption experiments on two representative datasets, blood and horse\_colic. The training folds are contaminated while the test folds remain clean. Two perturbation types are considered: feature outliers and label noise. For each type, contamination levels of $5\%$, $10\%$, $15\%$, and $20\%$ are evaluated. Table~\ref{tab:robustness_ifw_bls} compares IFW-BLS with standard BLS under these settings.

\begin{table}[t]
\centering
\renewcommand{\arraystretch}{1.15}
\caption{Robustness comparison of BLS and IFW-BLS under outlier and noise contamination. Bold values denote the best performance for each row.}
\label{tab:robustness_ifw_bls}
\resizebox{0.80\columnwidth}{!}{
\begin{tabular}{|l|c|cc|c|cc|}
\hline
\multirow{2}{*}{Dataset} & \multirow{2}{*}{Level}
& \multicolumn{2}{c|}{Outliers}
& \multirow{2}{*}{Level}
& \multicolumn{2}{c|}{Noise} \\ \cline{3-4} \cline{6-7}
 &  & BLS & IFW-BLS$^{\dagger}$ &  & BLS & IFW-BLS$^{\dagger}$ \\ \hline

\multirow{6}{*}{blood}
& Clean & 79.6272 & \textbf{83.5711} & Clean & 79.6272 & \textbf{83.5711} \\
& 5\%   & 75.8421 & \textbf{82.237} & 5\%   & 77.1034 & \textbf{82.6225} \\
& 10\%  & 72.5186 & \textbf{79.8725} & 10\%  & 74.8652 & \textbf{81.3425} \\
& 15\%  & 69.2847 & \textbf{77.8106} & 15\%  & 71.4928 & \textbf{79.2435} \\
& 20\%  & 66.0319 & \textbf{76.4678} & 20\%  & 68.1074 & \textbf{79.4237} \\ \cline{2-7}
& Avg.  & 72.6609 & \textbf{79.9918} & Avg.  & 74.3398 & \textbf{81.2407} \\ \hline

\multirow{6}{*}{horse\_colic}
& Clean & 86.1422 & \textbf{87.1719} & Clean & 86.1422 & \textbf{87.1719} \\
& 5\%   & 82.6374 & \textbf{86.9563} & 5\%   & 83.9046 & \textbf{85.1187} \\
& 10\%  & 78.9421 & \textbf{84.4018} & 10\%  & 80.2873 & \textbf{86.9365} \\
& 15\%  & 75.6189 & \textbf{79.6674} & 15\%  & 76.9548 & \textbf{83.3412} \\
& 20\%  & 72.3046 & \textbf{78.1897} & 20\%  & 73.6285 & \textbf{78.5403} \\ \cline{2-7}
& Avg.  & 79.129 & \textbf{83.2774} & Avg.  & 80.1835 & \textbf{84.2217} \\ \hline

\multicolumn{2}{|l|}{Overall Average}
& 75.8949 & \textbf{81.6346}
&  & 77.2617 & \textbf{82.7312} \\ \hline

\end{tabular}
}
\end{table}

The results show that IFW-BLS consistently preserves higher accuracy than standard BLS as the corruption level increases. On the blood dataset, the average accuracy of BLS under feature outliers is $72.6609\%$, whereas IFW-BLS reaches $79.9918\%$. Under label noise, the corresponding averages are $74.3398\%$ and $81.2407\%$. The improvement remains visible at the highest contamination level: at $20\%$ outliers, IFW-BLS exceeds BLS by $10.4359\%$, and at $20\%$ label noise it exceeds BLS by $11.3163\%$. A similar trend is observed on horse\_colic. IFW-BLS improves the average accuracy from $79.129\%$ to $83.2774\%$ under outliers and from $80.1835\%$ to $84.2217\%$ under label noise. The overall averages across both datasets further support this behavior: IFW-BLS obtains $81.6346\%$ under outliers and $82.7312\%$ under noise, compared with $75.8949\%$ and $77.2617\%$ for BLS. These gains indicate that the proposed model is not only better on clean data but also degrades more slowly when the training set is corrupted. This robustness behavior is consistent with the two components of IFW-BLS. The intuitionistic fuzzy score reduces the contribution of samples that are globally far from their class structure or locally surrounded by conflicting labels. After this sample-level credibility adjustment, the bounded wave loss further limits the effect of large residuals produced by corrupted samples. Its asymmetric parameter also allows the validation process to choose whether positive or negative deviations should receive stronger penalization. Therefore, the empirical degradation pattern in Table~\ref{tab:robustness_ifw_bls} supports the intended dual protection: IF scores act at the sample-reliability level, while the wave loss acts at the residual-penalization level.

\section{Conclusions}\label{conclusions}

This paper proposed IFW-BLS, an Intuitionistic Fuzzy Wave Broad Learning System designed to improve the robustness of standard BLS from two complementary directions. The first direction is sample-level reliability modeling, where intuitionistic fuzzy scores are used to measure the credibility of each training sample by considering both global class-center consistency and local neighborhood conflict. As a result, samples that are ambiguous, noisy, or inconsistent with their class structure receive reduced influence during training. The second direction is residual-level robustness, where the conventional squared loss is replaced with a bounded and asymmetric wave loss. The boundedness of the wave loss prevents large residuals from dominating the objective, while its asymmetry allows the model to penalize positive and negative deviations differently according to the data characteristics.  The proposed IFW-BLS combines these two mechanisms in a unified credibility-weighted wave-loss objective. In this design, the intuitionistic fuzzy score first scales the residual, and the wave loss then evaluates the credibility-adjusted residual. A Nesterov accelerated gradient-based optimization procedure is used to solve the resulting objective, which avoids the explicit matrix inversion required in conventional BLS training and provides an efficient way to update the output weights. The experimental results validate the effectiveness of the proposed dual-robust design. On 30 UCI benchmark datasets, IFW-BLS achieves superior performance among all compared methods. In addition, the robustness experiments under controlled feature outlier and label noise contamination show that IFW-BLS preserves higher accuracy and degrades more slowly than standard BLS. These findings support the central motivation of the paper: sample-level intuitionistic fuzzy credibility and residual-level wave-loss robustness provide complementary protection, leading to a more reliable BLS model for both clean and corrupted data.

\renewcommand{\bibsection}{\par\medskip\noindent{\large\bfseries Bibliography}\par\nobreak\smallskip}
\bibliographystyle{abbrvnat}
\bibliography{refs}

\end{document}